# Blind Deconvolution via Maximum Kurtosis Adaptive Filtering

Deborah Pereg

Doron Benzvi

The Jerusalem College of Engineering

Jerusalem, Israel

doronb@jce.ac.il , deborahpe@post.jce.ac.il

**ABSTRACT**

In this paper, we present an algorithm for identifying a parametrically described destructive unknown system based on a non-gaussianity measure. It is known that under certain conditions the output of a linear system is more gaussian than the input. Hence, an inverse filter is searched, such that its output is minimally gaussian. We use the kurtosis as a measure of the non-gaussianity of the signal. A maximum of the kurtosis as a function of the deconvolving filter coefficients is searched. The search is done iteratively using the gradient ascent algorithm, and the coefficients at the maximum point correspond to the inverse filter coefficients. This filter may be applied to the distorted signal to obtain the original undistorted signal. While a similar approach has been used before, it was always directed at a particular kind of a signal, commonly of impulsive characteristics. In this paper a successful attempt has been made to apply the algorithm to a wider range of signals, such as to process distorted audio signals and destructed images. This innovative implementation required the revelation of a way to preprocess the distorted signal at hand. The experimental results show very good performance in terms of recovering audio signals and blurred images, both for an FIR and IIR distorting filters.

# 1. Introduction

In signal processing blind deconvolution is an operation that, ideally, unravels the effect of a convolution performed by an unknown linear time-invariant system operating on an unknown input signal. Throughout the years, many different methods for blind deconvolution have been developed .Some of these methods were specifically adjusted for certain applications. But all of them have the same primary objective, which is to recover the input data signal from the faulty signal at hand.

In many applications, we can model the received signal (the observation) as

$$x(n) = s(n) * h(n) = \sum_{k=-\infty}^{\infty} h(k) s(n-k) \qquad (1)$$

where s(n) is the original source signal, h(n) is the impulse response of the destructive system and '*' denotes the one-dimensional linear convolution operation. The requirement is, therefore, to find the inverse system impulse response and reconstruct the input signal.

In this work, we investigate the use of blind deconvolution based on maximization of a non-gaussianity measure. Each sample of the observation signal is considered to be a random variable, generated from a linear combination of previous samples of the source signal. Assuming that the summed samples are statistically independent (which is not a trivial assumption) and that the original signal is not gaussian, it could be claimed that according to the Central Limit Theorem the probability density function (the pdf) of the observation signal is more gaussian than the pdf of the source signal. Thus, the source signal could be revealed by minimizing a gaussianity measure of the signal. This approach has already been used for blind deconvolution. However, its use was limited to the restoration of impulsive signals, such as seismic signals, speech de-reverberation [10] and the restoration of images by use of a genetic algorithm [7]-[9]. Here, an attempt has been made to apply the method to a general one-dimensional (audio) and two-dimensional (image) signals.

The remainder of the paper is organized as follows. In section 2.1 the kurtosis is reviewed and is related to the system identification problem. In section 2.2 the algorithm is described. In section 3, the implementation and the experimental results of our approach are reported. And lastly, section 4 includes some brief conclusions.

## 2. Maximun Kurtosis Adaptive Filtering

### 2.1 Kurtosis as a non-gaussianity measure

The kurtosis of a zero-mean random variable x is defined as its normalized fourth moment

$$k = \frac{E\{x^4\}}{\left[E\{x^2\}\right]^2} - 3 \qquad (2)$$

If x is a Gaussian random variable then $k = 0$. For this reason, the normal distribution is called mesokurtic. When $k > 0$ the distribution is called leptokurtic, and when $k < 0$ the distribution is called platykurtic.

**Fig.1** shows the histogram of the original signal which is an i.i.d input signal $s(n)$ with uniformly distributed and the histogram of $x(n)$ the output of an IIR system:

$$x(n) = a_1 x(n-1) + a_2 x(n-1) + s(n)$$

The output of the system is more gaussian and its kurtosis is closer to zero. Thus, the inverse filter, restoring the source signal, generates a signal which is less gaussian, maximizing the kurtosis of the signal.

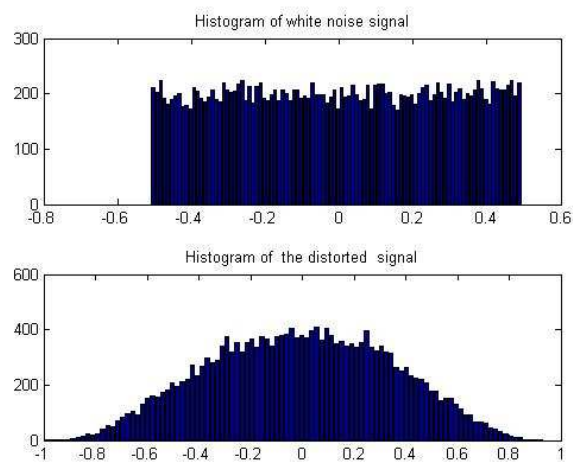

**Fig. 1** Histograms of the input and the output of an LTI system. The output is more Gaussian.

Consequently, in order to find the correct inverse system and apply it to the observation signal x(n), we wish to develop an on-line adaptive gradient-ascent algorithm that maximizes the kurtosis' (of the whitened signal) absolute value.

When the original signal is not i.i.d, it has to be whitened to comply with the central limit theorem. Since the source signal is not available, whitening is applied to the observation signal before we try to maximize its non-gaussianity. This change of order of operations is acceptable for the LTI systems involved. Whitening can be done before or after the deforming system. In other words, our goal is to find the blind deconvolution filter that makes the whitened observation signal as far as possible from being gaussian.

To test this concept, we performed the following experiment: We used an IIR system with 2 parameters to destruct an input signal s(n). Mathematically, we created an output signal

$$x(n) = a_1 x(n-1) + a_2 x(n-1) + s(n)$$

Then, we whitened x(n) by use of a highpass filter or Linear Predictive Code (LPC). Let us use $x_1(n)$ to denote the whitened signal. We applied the inverse FIR system

$$s_{est}(n) = x_1(n) - a_{1,est} x_1(n-1) - a_{2,est} x_1(n-2)$$

and calculated the kurtosis' absolute value of $s_{est}(n)$ for different values of the parameters $a_{1,est}, a_{2,est}$ ($-1 < a_{1,est}, a_{2,est} < 1$). The maximum value of $kurt\{s_{est}(n)\}$ is achieved in most of the tests for the correct inverse system, i.e. the system with the correct parameters:

$$a_{1,est} \cong a_1, \quad a_{2,est} \cong a_2$$

Fig. 2 is an example of this experiment. The parameters chosen were $a_1 = 0.6$ and $a_2 = -0.3$. s(n), the original signal, was a speech signal. $x_1(n)$ was obtained by use of LPC. The maximum value of $kurt\{s_{est}(n)\}$ was achieved for $a_{1,est} \cong a_1 = 0.62$, $a_{2,est} \cong a_2 = -0.28$.

Rather than performing an exhaustive search for the correct parameters of the inverse filter, those that maximize the kurtosis of the output, an iterative search for the maximum point may be done using the gradient ascent algorithm.

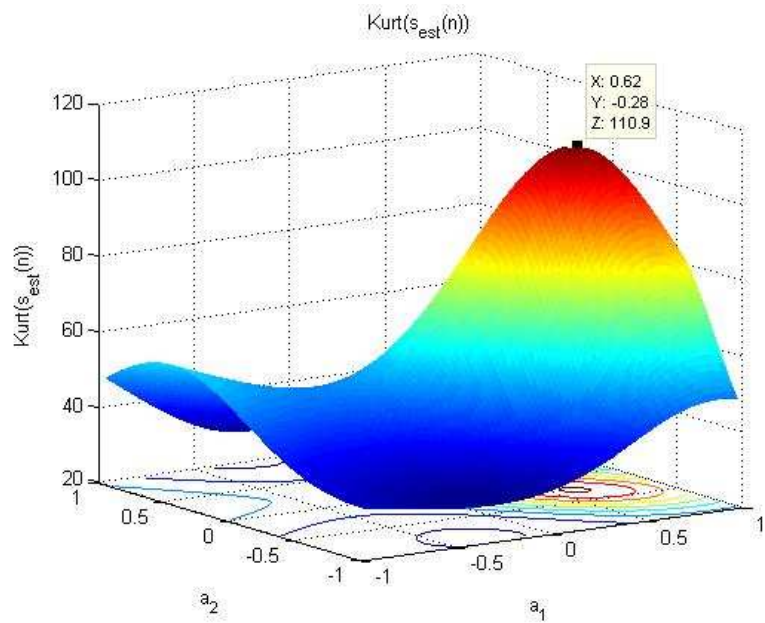

**Fig.2** The kurtosis of $s_{est}(n) = x_1(n) - a_{1,est} x_1(n-1) - a_{2,est} x_1(n-2)$

for $-1 < a_{1\,est}, a_{2,est} < 1$. The maximum is achieved when $a_{1,est} \cong a_1,\ a_{2,est} \cong a_2$

### 2.2 The algorithm

The correct coefficients of the inverse filter are achieved using the gradient ascent algorithm, as mentioned above. The normalized kurtosis is the value to be maximized. The filter is initialized to an all-pass filter. On each iteration its coefficients are updated in the direction of the gradient, which is the direction of the greatest rate of increment/decrement of the value of the cost function – the normalized kurtosis.

Fig.3 shows the blind-deconvolution system. The received, destructed observation audio\image signal is x(n). Its corresponding whitened signal is $x_1(n)$. **h**(n) is the L-tap adaptive filter at time n. The output is derived by

$$\tilde{y}(n) = h^t(n)\tilde{x}_1(n) \qquad (3)$$

where

$$\tilde{x}_1(n) = \left[ x_1(n-L+1)\ldots x_1(n-1)\, x_1(n) \right]^T$$

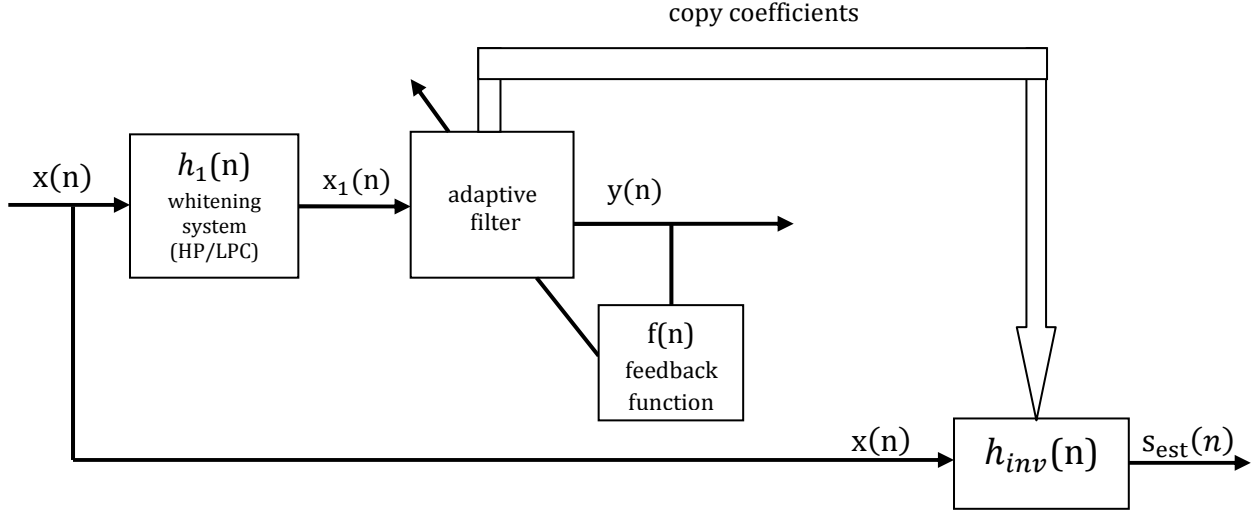

**Fig.3** An on-line adaptive time-domain adaptive algorithm for maximization of the non-gaussianity of the whitened signal.

The whitening system can be an LP synthesis filter when x(n) can be modeled as an AR process. However, in most cases the use of a highpass filter for whitening is sufficient. The adaptation of **h**(n) is similar to the traditional LMS adaptive filter [2], with a feedback function f(n), derived from the gradient of the kurtosis, as in [10]. To derive the following adaptation equations, recall that we are looking for a filter that maximizes or minimizes the kurtosis of $\tilde{y}(n)$ which is given by

$$J(n) = \frac{E\{y^4(n)\}}{E\{y^2(n)\}} - 3 \qquad (4)$$

where the expectation $E\{\cdot\}$ can be estimated from the samples' average. The gradient of $J(n)$ with respect to current filter is

$$\frac{\partial J}{\partial \boldsymbol{h}} = \frac{4[E\{\tilde{y}^2\} E\{\tilde{y}^3 \tilde{\boldsymbol{x}}_1\} - E\{\tilde{y}^4\} E\{\tilde{y}^4 \tilde{\boldsymbol{x}}_1\}]}{E^3\{\tilde{y}^2\}} \qquad (5)$$

The dependence on the time n is not written for simplicity. In a manner similar to [10] we can approximate the gradient by

$$\frac{\partial J}{\partial \boldsymbol{h}} = \left( \frac{4\left[\left(E\{\tilde{y}^2\}\tilde{y}^2 - E\{\tilde{y}^4\}\right)\tilde{y}(n)\right]}{E^3\{\tilde{y}^2\}} \right) \tilde{\boldsymbol{x}}_1(n) = f(n)\tilde{\boldsymbol{x}}_1(n) \qquad (6)$$

We refer to f(n) as the feedback function. This function is used to control the filter updates. The final structure of the update equation for the filter that maximizes the absolute value of the kurtosis of $x_1(n)$ is then given by

$$\boldsymbol{h}(n+1) = \boldsymbol{h}(n) + \mu f(n)\tilde{\boldsymbol{x}}_1(n) \qquad (7)$$

where

$$f(n) = \frac{4\left[\left(E\{\tilde{y}^2\}\tilde{y}^2 - E\{\tilde{y}^4\}\right)\tilde{y}(n)\right]}{E^3\{\tilde{y}^2\}}$$

The second and fourth moments of the observation signal are calculated recursively:

$$E\{\tilde{y}^2(n)\} = \beta E\{\tilde{y}^2(n-1)\} + (1-\beta)\tilde{y}^2(n)$$

and

$$E\{\tilde{y}^4(n)\} = \beta E\{\tilde{y}^4(n-1)\} + (1-\beta)\tilde{y}^4(n)$$

The parameter μ controls the speed of adaptation. The sign of μ is determined according to the nature of the source signal. For a super-gaussian s(n), μ > 0. For a sub-gaussian s(n), μ < 0. We often have some a priori information about the nature of s(n). For example, speech signals are usually highly super-gaussian, and image signals are usually sub-gaussian. The parameter β controls the smoothness of the moment estimates. The value of the β coefficient is chosen to be 0.99 in most cases.

When the signal is a 2-D signal (an image) there are two options for implementing the suggested algorithm. One way is to convert the image into a 1-D vector by chaining its rows. In this way, the inverse vector filter will be very long and most of its values should be forced to zero on each iteration of the algorithm. Alternatively, the algorithm can be implemented with images and a 2-D inverse filter of size MxN. Then, $\tilde{x}_1(n,m)$ is defined as

$$\tilde{x}_1(n,m) = \begin{pmatrix} \tilde{x}_1\left(n-\frac{N-1}{2}, m-\frac{M-1}{2}\right) & \cdots & \tilde{x}_1\left(n-\frac{N-1}{2}, m+\frac{M-1}{2}\right) \\ \vdots & \tilde{x}_1(n,m) & \vdots \\ \tilde{x}_1\left(n+\frac{N-1}{2}, m-\frac{M-1}{2}\right) & \cdots & \tilde{x}_1\left(n+\frac{N-1}{2}, m+\frac{M-1}{2}\right) \end{pmatrix}$$

And similarly the output is

$$\tilde{y}(n,m) = h^t(n,m)\tilde{x}_1(n,m)$$

In the restoration of images whitening is performed only by creating a differential image, which can be done by using a simple highpass filter.

## 3. Experimental Results

In this section experimental results obtained from applying the proposed algorithm to example signals are presented. To measure the accuracy of the system identification, we measure the normalized correlation between the original and the reconstructed signal, and compare it to the correlation between the original signal and the distorted signal.

### 3.1 Reverberating speech signals

The proposed scheme was first tested on speech signals degraded by an IIR system mathematically described by

$$x(n) = a_1 x(n-D) + a_2 x(n-2D) + s(n)$$

The parameters $a_1, a_2$ are chosen so that the system is stable. $D \gg 1$, so that the system creates a noticeable echo effect. Since $D \gg 1$, the system generates a linear combination of samples, which are far apart, and thus can be considered statistically independent. Therefore whitening is not necessary in this case.

Fig. 4(a) shows segments from two original speech waves; Fig 4(b) shows the corresponding reverberating speech signals generated by applying the system above to the original signal. Fig. 4(c) presents the restored signals. A delay value of D=100 is used. The optimal inverse filter parameters, the correlation coefficients between the original signal and the distorted signal and between the original signal and the estimated signal ($\rho_{sx}$ and $\rho_{s\hat{s}}$ respectively) and the kurtosis values of the source, observation and restored signals are summarized in Table 1.

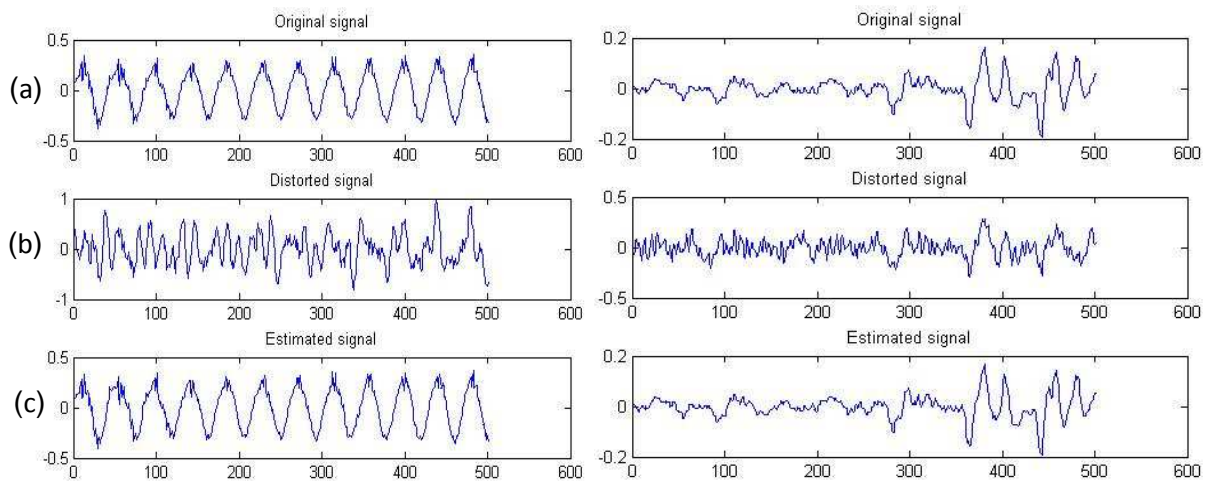

**Fig 4.** Blind Deconvolution results for reverberating speech signals.

**Table 1.** Parameters and correlation performance of tested restoration of reverberating speech signals.

|  | $a_1$ | $a_{1,est}$ | $a_2$ | $a_{2,est}$ | $|kurt(s)|$ | $|kurt(x)|$ | $|kurt(s_{est})|$ | $\rho_{sx}$ | $\rho_{s\hat{s}}$ |
|---|---|---|---|---|---|---|---|---|---|
| Left signal | -0.6 | -0.54 | 0.3 | 0.4 | 3.163 | 0.690 | 3.146 | 0.489 | 0.988 |
| Right signal | -0.5 | -0.48 | 0.4 | 0.42 | 17.481 | 3.621 | 17.668 | 0.502 | 0.999 |

## 3.2 Audio signals restoration

In the first part of this experiment, an IIR filter is used to degrade a speech signal. The degrading system we used was

$$x(n) = a_1 x(n-1) + a_2 x(n-2) + s(n)$$

Fig. 5 shows the result s of this experiment for 4 different speech segments. The signals in Fig. 5 (a) and (b) were whitened by use of a highpass filter, while the signal in Fig. 5(c) and (d) were whitened by use of LPC (of 5'th order ). Table 4 compares the estimated parameters to the correct parameters and the correlation coefficients.

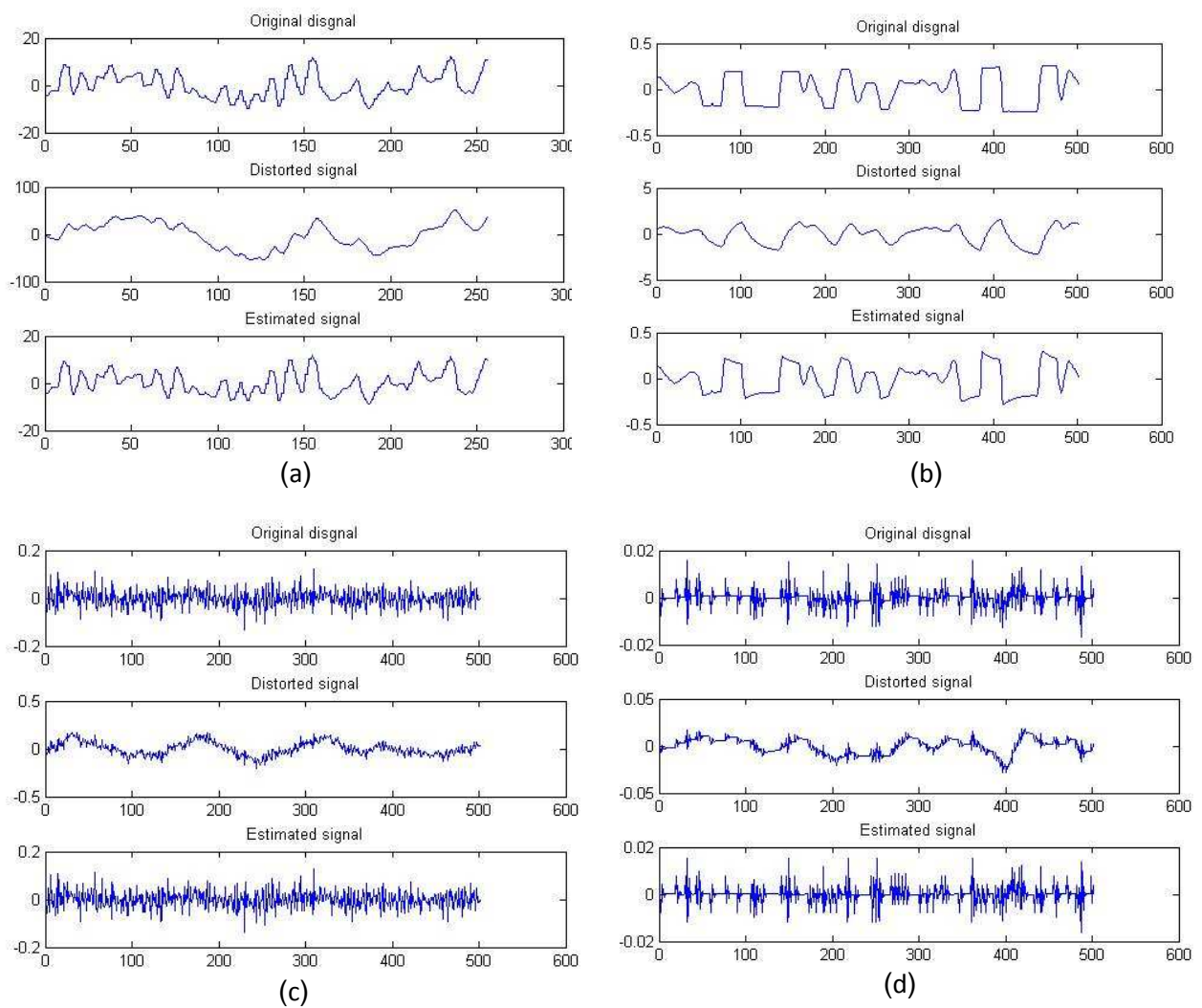

**Fig.5** Blind Deconvolution of IIR degraded audio signals.

**Table 2.** Parameters and correlation performance of tested restoration of distorted 1-D signals.

|  | $a_1$ | $a_{1,est}$ | $a_2$ | $a_{2,est}$ | $\rho_{sx}$ | $\rho_{s\hat{s}}$ |
|---|---|---|---|---|---|---|
| (a) Whitening with HP filter Heart rate signal | 0.6 | 0.601 | 0.3 | 0.344 | 0.636 | 0.972 |
| (b) Whitening with HP filter Speech signal | 0.5 | 0.466 | 0.4 | 0.461 | 0.633 | 0.989 |
| (c) Whitening by LPC Speech and music signal | 0.5 | 0.555 | 0.4 | 0.352 | 0.534 | 0.998 |
| (d) Whitening by LPC Speech and music signal | 0.6 | 0.639 | 0.3 | 0.341 | 0.483 | 0.986 |

In the second part of this experiment, an FIR filter is used to degrade a speech signal. The degrading system in use was

$$x(n) = (a_1 + a_2) * s(n-1) + (a_1 a_2) * s(n-2) + s(n)$$

The IIR inverse filter is as approximated by a long FIR filter.

Theoretically, the first 3 coefficients of the inverse filter are:

$$h(0) = 1, h(1) = -(a_1 + a_2), h(2) = a_1^2 + a_1 a_2 + a_2^2$$

Fig. 6 shows the results of this experiment for 3 different speech segments. The signals in Fig. 6 (b) and (c) were whitened by use of a highpass filter, while the source signal in Fig. 6(a) was already white, therefore whitening was not necessary. Table 3 compares the estimated parameters to the correct parameters and the correlation coefficients.

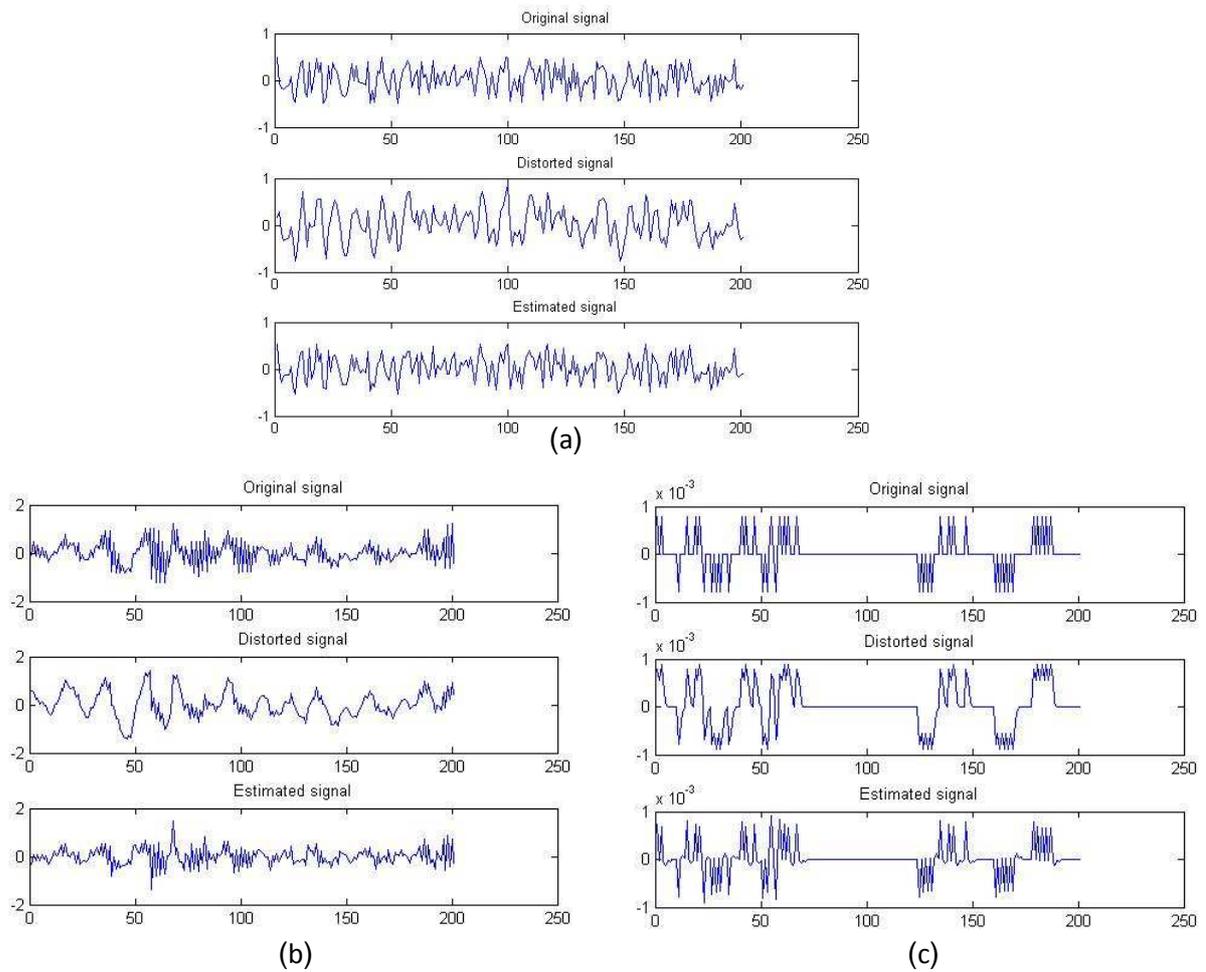

**Fig.6** Blind Deconvolution of FIR degraded audio signals

**Table 3.** Parameters and correlation performance of tested restoration of distorted 1-D signals

|  | $a_1 + a_2$ | $h(1)$ | $a_1^2 + a_1 a_2 + a_2^2$ | $h(2)$ | $\rho_{sx}$ | $\rho_{s\hat{s}}$ |
|---|---|---|---|---|---|---|
| (a) White source signal (no whitening) | 0.8 | -0.701 | 0.49 | 0.306 | 0.778 | 0.98 |
| (b) Whitening with HP filter Heart rate signal | 0.8 | -0.815 | ---- | ---- | 0.798 | 0.912 |
| (c) Whitening with HP filter Speech signal | 0.7 | -0.707 | 0.37 | 0.262 | 0.829 | 0.99 |

## 3.3 Distorted Images

The proposed scheme was tested on various images degraded by an IIR system with 2 to 3 parameters. Mathematically the destructing system in this experiment can be the described by the difference equation

$$g(x,y) = \sum_{k=-M}^{k=M} \sum_{l=-N}^{l=N} a_{kl} g(x-k, y-l) + f(x,y)$$

where g(x,y) is the blurred image, f(x,y) is the original image and $a_{kl}$ are the parameters that will be optimized in search of the de-blurring system.

Whitening in this case is done by high-pass filtering, generating the 2D difference image.

Fig 6. presents the restoration of 4 test images distorted by use of the IIR system:

$$g(x,y) = a_1 g(x-1, y) + a_2 g(x, y-1) + f(x,y)$$

Theoretically the correct inverse filter in this case is:

$$w = \begin{pmatrix} 0 & -a_1 & 0 \\ -a_2 & 1 & 0 \\ 0 & 0 & 0 \end{pmatrix}$$

Our algorithm was able to estimate a filter with approximately the correct parameters, and the restoration was successful. Table 2 shows the identified parameters and the correlation coefficients $\rho_{sx}$ and $\rho_{s\hat{s}}$ The results shows that the algorithm can give a good restoration without knowing the blurring process.

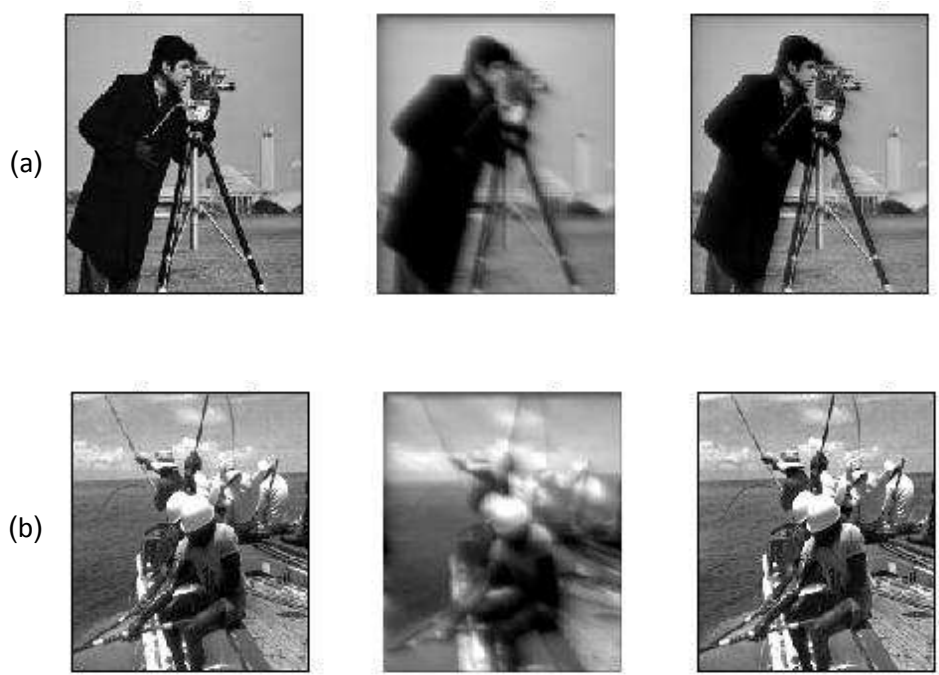

(a)

(b)

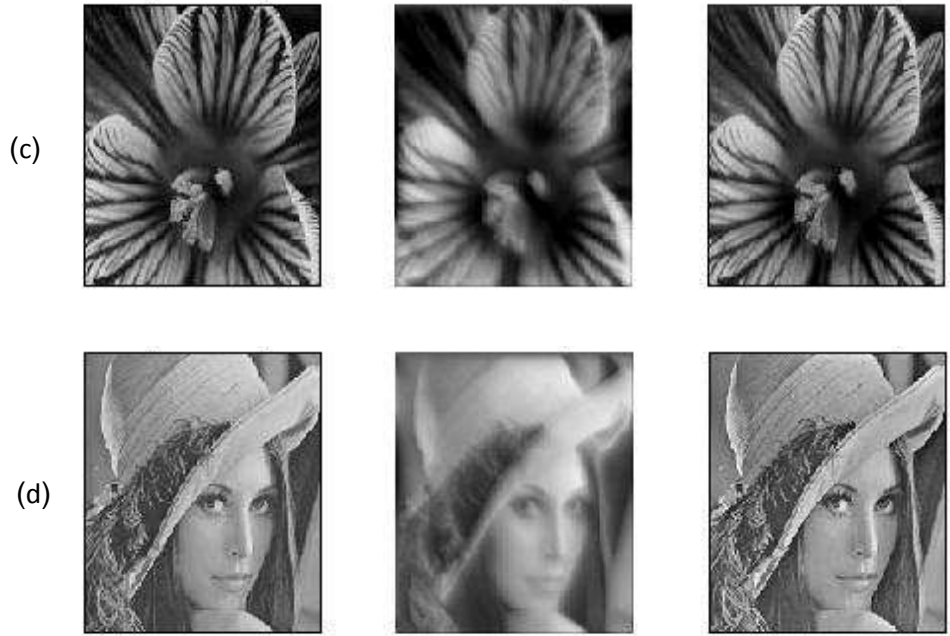

**Fig 7**. Blind deconvolution of blurred images. column 1:original images, column 2: blurred images, column 3: restored images.

**Table 4**. Parameters and correlation performance of tested restoration of distorted images

|     | $a_1$ | $a_{1,est}$ | $a_2$ | $a_{2,est}$ | $\rho_{sx}$ | $\rho_{s\hat{s}}$ |
|-----|-------|-------------|-------|-------------|-------------|-------------------|
| (a) | 0.5   | 0.515       | 0.4   | 0.4044      | 0.847       | 0.972             |
| (b) | 0.5   | 0.494       | 0.4   | 0.401       | 0.816       | 0.987             |
| (c) | 0.6   | 0.618       | 0.3   | 0.281       | 0.766       | 0.983             |
| (d) | 0.6   | 0.524       | 0.3   | 0.304       | 0.779       | 0.989             |

Fig 8. shows the restoration of 2 images distorted by the IIR system

$$g(x,y) = a_1 g(x-1,y) + a_2 g(x,y-1) + a_3 g(x-1,y-1) + f(x,y)$$

Almost the exact parameters were identified using the proposed algorithm.

Table 3 presents these parameters and the correlation coefficients' values.

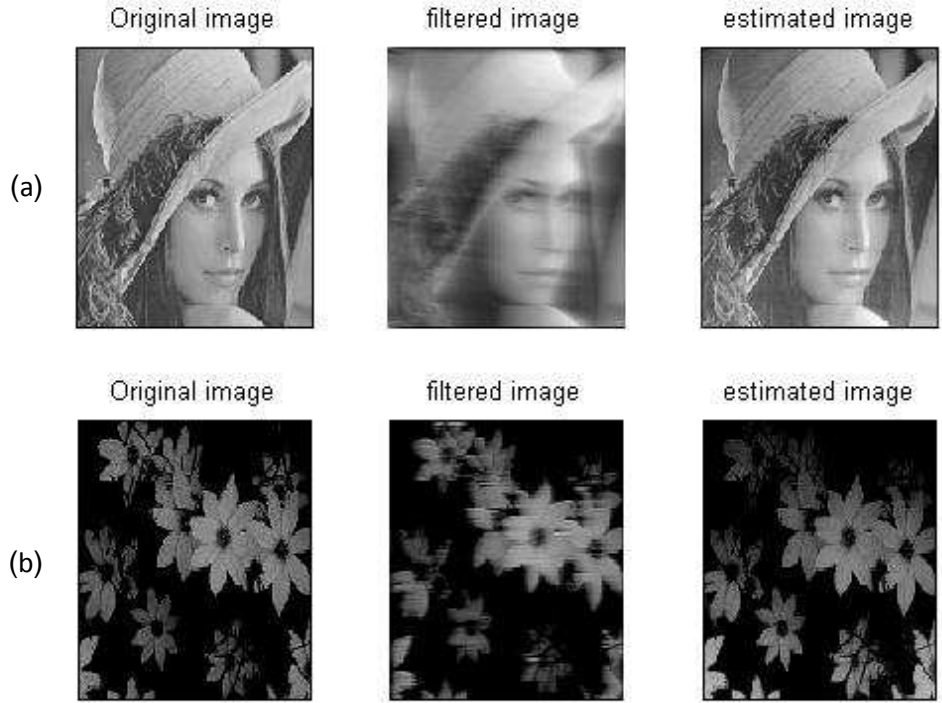

**Fig 8.** . Blind deconvolution of blurred images . column 1:original images, column 2: distorted images, column 3: restored images.

Table 5. Parameters and correlation performance of tested restoration distorted images

|     | $a_1$ | $a_{1,est}$ | $a_2$ | $a_{2,est}$ | $a_3$ | $a_{3,est}$ | $\rho_{sx}$ | $\rho_{s\hat{s}}$ |
|-----|-------|-------------|-------|-------------|-------|-------------|-------------|-------------------|
| (a) | 0.8   | 0.75        | -0.3  | -0.35       | 0.2   | 0.25        | 0.701       | 0.977             |
| (b) | 0.8   | 0.707       | -0.4  | -0.404      | 0.5   | 0.543       | 0.870       | 0.998             |

The proposed method has been implemented when the destructing system is and IIR or FIR system with more than 2 or 3 parameters. For simplicity, we chose to demonstrate IIR systems with up to 3 parameters, so that the inverse system is a FIR system and the requirement is to identify only up to 3 parameters. When the deforming system is of type FIR, the inverse filter is of type IIR. However, the algorithm is employed while approximating the inverse filter to be of a finite length. One of the limitations of our algorithm is the need for a priori knowledge of the number of parameters to be estimated. In other word, the length of the inverse system is to be known. This limitation does not exist when the original signal is an i.i.d signal.

## 4. Conclusions

In this paper we proposed an adaptive filtering method based on non-gaussianity measure (kurtosis), to restore reverberating speech signals, distorted images and degraded audio signals. We validated the performance of our algorithm on artificially difficult reverberating speech, distorted images and degraded audio signals. The results indicate that the proposed method is able to restore 1-D and 2-D signals deformed by both IIR and FIR systems. Future endeavors are to investigate the use of the algorithm for restoration of naturally distorted signals, and to compare the algorithm to a similar algorithm based on other non-gaussianity measures such as Negentropy.